# AsNER - Annotated Dataset and Baseline for Assamese Named Entity recognition

**Dhrubajyoti Pathak, Sukumar Nandi, Priyankoo Sarmah**
Indian Institute of Technology Guwahati
North Guwahati, India
{drbj153, sukumar, priyankoo}@iitg.ac.in

**Abstract**

We present the AsNER, a named entity annotation dataset for low resource Assamese language with a baseline Assamese NER model. The dataset contains about 99k tokens comprised of text from the speech of the Prime Minister of India and Assamese play. It also contains person names, location names and addresses. The proposed NER dataset is likely to be a significant resource for deep neural based Assamese language processing. We benchmark the dataset by training NER models and evaluating using state-of-the-art architectures for supervised named entity recognition (NER) such as Fasttext, BERT, XLM-R, FLAIR, MuRIL etc. We implement several baseline approaches with state-of-the-art sequence tagging Bi-LSTM-CRF architecture. The highest F1-score among all baselines achieves an accuracy of 80.69% when using MuRIL as a word embedding method. The annotated dataset and the top performing model are made publicly available.

## 1. Introduction

Named Entity Recognition (NER) aims to classify text in a sentence into predefined classes such as person, location, organization etc. It basically identifies a word or phrase that can be considered as names from a set of documents. NER plays a vital role in preprocessing task of various natural language processing (NLP) applications such as information retrieval (Neudecker, 2016), text understanding (Zhang et al., 2019), automatic text summarization (Larsen, 1999), question answering (Mollá et al., 2006), machine translation (Babych and Hartley, 2003), and knowledge base construction (Etzioni et al., 2005) etc. Recent advances in deep neural network (DL) in NLP exhibit success in various domains. DL-based NER systems with minimal feature engineering have been flourishing. Over the past few years, several studies have been reported success in deep learning-based NER model and achieved state-of-the-art performance in resource-rich language (Chiu and Nichols, 2016; Lample et al., 2016; Akbik et al., 2019). The DL-based model requires high quality and large annotated datasets for training and evaluation. Therefore, datasets play an essential role in extracting the linguistic features of a language to achieve high performance in downstream tasks. On the other hand, data annotation for a language remains a time consuming and expensive process. It is a major challenge for many resource-poor languages, such as Assamese (Glottocode: *assa1263*), as it requires language experts to perform annotation tasks on a large amount of data. Suitable annotated corpora for Assamese NER, name dictionaries, morphological analyzers, dependency parser, POS taggers etc., are not yet publicly available for suitable use in downstream tasks. Although, Assamese has a very old and rich literary history, technology development in NLP is still in a nascent stage. The WikiAnn NER dataset (Pan et al., 2017) is the only publicly available dataset that contains annotated NER data for 282 languages that exist in Wikipedia; however, the size is not large enough to train a neural model. The lack of a suitable dataset may be the reason that we could not find any study about DL-based NER for Assamese. All the previous studies on Assamese NER are based on rule-based or statistical approaches.

In this paper, we introduce a novel Assamese NER dataset, AsNER, comprising of annotated sentences with five entity classes. It also contains a large number of person names, locations and organization names. The corpus is built from the Assamese translation of the speeches of the Prime Minister of India, available online. Along with the annotation, preprocessing of the dataset and manual evaluation of the proposed AsNER, we verify the effectiveness of our dataset by using it to train neural-based NER models. To the best of our knowledge, AsNER is the first attempt to develop and evaluate NER dataset for Assamese language. The summary of our contribution is as follows:

- We prepare and release a novel NER dataset for low-resourced Assamese language.

- We present an evaluation of the AsNER by employing a state-of-the-art sequence tagging BiLSTM-CRF architecture.
- Lastly, we report the performance of eight different NER models trained on AsNER.
- The dataset and the best performing trained model are made available publicly[1].

This paper is organized as follows. We present a brief overview of the Assamese language in section 2. We describe our annotated dataset and annotation process in Section 3 and 4 respectively. We illustrate experiment details used to evaluate our dataset in Section 5. We discuss the results of the performance of different models in Section 6. In 7, we report the challenges encountered in the annotation process as well as in training the models. Finally, we conclude our paper in section 8.

## 2. Assamese language

Assamese or Asamiya, pronounced, /ɔxɔmijɑ/ is an Indo-Aryan language spoken mainly in Assam, a state of northeast India. It also refers to the native of Assam, whose mother tongue is Assamese. In this work, through Assamese, we refer to the Assamese language. It is a descendant of Magadhi Prakrit and bears affinities with Bengali, Hindi and Odia. Modern Assamese uses the Assamese script, which is developed from the Brahmi script. Assamese script is similar to Bengali script except for two characters, where Assamese differing from Bengali in one letter (ৰ) for the /r/ sound, and an extra letter (ৱ) for the /w/ or /v/ sound.

Assamese is a highly inflectional, morphologically rich and agglutinating language. The rich linguistics feature of the language becomes the most challenging tasks in language processing. The morphological structure of words changes due to affixation, derivation and compounding. Affixes play an important role in word formation in Assamese. Affixes are used extensively in the formation of nouns, pronouns, and in the inflection of verbs with respect to number, person, tense, aspect and mood. Assamese is also a free word order Language.

## 3. Dataset description

There are a few organized sources of monolingual corpora available for most of the Indian languages. Among the Indian languages, the Assamese corpus is one of the smallest in size. The WikiAnn NER[2] dataset is the only publicly available dataset for Assamese NER. It is created from Wikipedia by transferring named entity labels from English to other languages by utilizing cross language link and Knowledge Base properties. The size of WikiAnn is not too large to train a neural model. The publicly available Assamese NER comprises of 12.5k tokens.

The major part of the dataset is taken from the translated speech of the Prime Minister published by Press Information Bureau, Govt. of India[3]. Text from Assamese Wikisource[4] and Wikipedia [5] is also included. Apart from that we include person names, location names and organization names into the dataset. These persons, locations and organization's names, are originally in English text collected from various sources, and translated to Assamese using Microsoft translator[6]. The statistics of sentence and entity count from all the sources is presented in table 1. The final dataset comprises 24,040 sentences and approximately 99k tokens. We used 80% of the dataset for training, while approximately 10% of dataset is used for validation and 10% for testing. The additional person, location and organization names are proportionally distributed in the training, validation and test sets. We summarise the statistics of the dataset in table 4.

## 4. Dataset Annotation

The annotated dataset is prepared in two phase; first by a POS tagger to tag all the words in the sentence with POS tag. After that, we use the tagged sentences for further named entity tagging by three native Assamese speakers with one linguist annotator. In second phase, only the nouns and the numbers are kept for further checking. The rest of the words of the sentence are annotated as "not a named entity". Table 3 shows the distribution of different classes in the dataset. Conflicts have been resolved manually. We discuss the conflicting cases in section 7

We focus on five classes of named entity during annotation, Location(LOC), Person (PER), Organization (ORG), Miscellaneous (MISC) and Number (NUM) (Sang and De Meulder, 2003; Chinchor and Robinson, 1997; Strötgen and Gertz, 2013; Hvingelby et al., 2020).

**LOC** includes locations like regions (villages, towns, cities), roads ( street name, highway), and natural locations (National park, forest reserve, river, garden), as well as both public and commercial places like tourist sites, museums, hospitals, airports, stations, markets, play-grounds, restaurants, hotels etc.

**PER** consists of first, middle and last names of people, animals, fictional characters, as well as

---

[1] https://anonymous.4open.science/r/AsNER-04B3/

[2] https://elisa-ie.github.io/wikiann/

[3] https://pib.gov.in/indexd.aspx/

[4] https://as.wikisource.org/

[5] https://as.wikipedia.org/

[6] https://www.microsoft.com/en-us/translator/

Table 1: Statistic of various source

| Corpus Source | #Sentence | #Entities |
|---|---|---|
| PM Speech text | 4534 | 6060 |
| Assamese play(Wiki) | 403 | 595 |
| Location text | 9769 | 16067 |
| Person text | 8798 | 9776 |
| Organisation text | 518 | 2465 |

Table 2: Dataset Statistics

| **Dataset** | **Train** | **Dev** | **Test** |
|---|---|---|---|
| # sentences | 21475 | 767 | 1798 |
| # tokens | 81422 | 8292 | 8909 |
| # entities | 29854 | 1326 | 3783 |

Table 3: Frequency distribution of the NER taggset in the dataset

| Class Name | Token count | | |
|---|---|---|---|
| | Train | Dev | Test |
| LOC | 16688 | 247 | 752 |
| PER | 10224 | 370 | 728 |
| ORG | 712 | 327 | 2071 |
| MISC | 1574 | 276 | 184 |
| NUM | 656 | 106 | 48 |
| O | 51568 | 6966 | 5126 |

aliases.
**ORG** includes public organisations (schools, colleges, universities, charities), companies ( Banks, stock, markets, company name, brands), government bodies ( ministries, institutions, courts, political unions of countries such as UNESCO).
**MISC** It includes a broad category such as nationalities, languages, political ideologies, religions, events (conferences, seminar, festivals, book fair, expo, sports competitions, forums, parties, concerts) etc. It also categorises words of which one part is a location, person or organisation, or other words derived from a word which is a location, organisation, or a person.
**NUM** includes numbers, money, percentage and quantity etc.
The tag **"O"** is used for the remaining tokens.
We measure the inter-annotator agreement using Cohen's kappa ($\kappa$). We observed inter-annotator agreement on 0.85 using $\kappa$.
The annotated dataset is prepared in column format in which each line represent a word, and each column represents one level of linguistic annotation. An empty line separate sentences in the dataset. e.g.

| দিল্লী | LOC |
| ভাৰতৰ | LOC |
| ৰাজধানী | O |
| । | O |
| দেৱজীত | PER |
| বৰুৱা | PER |
| দিল্লীত | LOC |
| থাকে | O |
| । | O |

The first column of the dataset is the word itself and the second one is the NER tag.

### 4.1. Resolving annotation conflict

We resolve the annotation conflict manually by following semantics and grammatical rules. Homonym ambiguity is prevalent in Assamese. Homonym ambiguity is resolved by looking at the context of the sentence. e.g. জোন /zʊn/ (Moon) would typically be a MISC, but when it is also often used in the name of a person in Assamese. In that case, it would be PER. Similarly, when an institution's name comprises a place name, it can act both ORG and LOC. e.g. আইআইটি গুৱাহাটী /iit guwahati/ is an educational institute which generally implies as an organization (ORG); however, in some instance, it can be referred to as a location (LOC) also. Therefore, it is a context-dependent whether to annotate as ORG or LOC.

## 5. Experiments

We choose a state-of-the-art neural sequence tagger framework based on standard BiLSTM-CRF architecture (Yu et al., 2020; Akbik et al., 2019; Huang et al., 2015) in our evaluation. It allows to apply various state-of-the-art language models to train sequence tagging models such as Named Entity Recognition (NER), and POS tagging, especially for high resource languages such as English, German and Dutch (Akbik et al., 2018; Peters et al., 2018). Using the architecture, we evaluate our dataset using different pre-trained word embeddings.

## 5.1. Word Embeddings used in evaluation

Word embedding is a crucial component in machine learning-based NLP models. The real-valued vector representation of words has the ability to capture both semantic and syntactic meanings of words in a sentence, which led to significant advances in recent natural language processing (NLP) tasks such as sequence classification, POS tagging, Named Entity Recognition (NER), Sentiment Analysis, Machine Translation (MT), Question Answering (QA). There are various word embeddings to embed the words in sentences in multiple ways. These embeddings are usually trained and evaluated on high-resource languages, using a large collection of unlabeled corpora to build feature-rich word embeddings. Eventually, it becomes an integral part of neural models in NLP applications and is found to be achieved state-of-the-art performance in the downstream task. We used eight different pre-trained word embeddings in our experiment of developing Assamese NER. In the next part, we briefly describe all these word embeddings.

**Word2Vec** (Mikolov et al., 2013; Kakwani et al., 2020): It is capable of capturing the context of a word in a sentence, semantic and syntactic similarity between the words. In Word2vec, word embeddings can be obtained by utilizing either of the two architectures, continuous bag-of-words (CBOW) or continuous skip-gram. We use the CBOW model in our experiment.

**FastText embedding** (Bojanowski et al., 2017) are belongs to the type of sub-word embeddings where it is trained on character $n$-grams of words rather than whole words. FastText Embeddings can give word vectors for out of vocabulary(OOV) words by using the sub-word information from the previously trained model.

**Byte-Pair Embeddings** (Heinzerling and Strube, 2018) are pre-computed on sub-word level. They can embed by splitting words into subwords or character sequences, looking up the pre-computed subwords embeddings.

**Bidirectional Encoder Representations from Transformers (BERT)** (Devlin et al., 2018) uses masked language models (MLM). In masked language model, one or more words of the input strings are randomly masked and tries to predict the masked word based on its context.

**Embeddings from Language Models (ELMo)** (Peters et al., 2018) word representation models complex features of words uses such as syntax and semantics. An embedding vector of a word varies according to the context of the sentences (i.e. to model polysemy). ELMo word vector representations are calculated based on the entire input sentence.

**Flair Embeddings** (Akbik et al., 2018) is a contextual string embeddings for any string of characters in a sentential context. The embedding method based on recent advances in neural language modelling (LM) (Sutskever et al., 2014) that provides languages to be modelled as distributions over sequences of characters instead of words.

**Multilingual Representations for Indian Languages (MuRIL)** (Khanuja et al., 2021) is based on BERT base architecture pre-trained on 17 Indian languages, including Assamese.

**XLM-R** (Conneau et al., 2019) uses self-supervised training techniques in cross-lingual understanding. In cross-lingual understanding, a model is trained for a task in one language and then used with other languages without additional training data.

Most of these word embeddings are trained on Assamese Wikipedia dataset (approx 8 million words) which size is limited for quality training. So, the performance in NER tagging is lower than most recently proposed MuRIL (Khanuja et al., 2021) which are trained on comparatively large size of training datasets.

## 5.2. Hyperparameters

We train our models on Nvidia Tesla P100 403 GPU (3584 Cuda Cores). We choose similar hyperparameters in all the eight experiments in table 4 to evaluate the performance of different pre-trained Assamese word embeddings over the AsNER dataset. We kept the hidden layer size of 1024 with six hidden layers and a maximum sequence length of 128. To account for memory constraints, we use a mini-batch size of 32. Apart from hidden size, hidden layer, sequence length and batch size, we use the default values for the remaining hyperparameters as in (Akbik et al., 2019). We train the model for 50 epochs. We use learning rate annealing for early stopping.

## 6. Results

We present the performance results of different experiments using our annotated dataset in this section. To conduct the experiments, we configure eight setups using different pre-trained word embedding models discussed in section 5.1. We use

Table 4: Summary of the performance of individual word-embeddings in sequence labelling task

| Experiment | Embeddings | NER Accuracy (F1 Score) |
|---|---|---|
| 1 | Word2Vec | 60.23% |
| 2 | FastTextEmbeddings | 67.94% |
| 3 | BytePairEmbeddings | 76.24% |
| 4 | BERT Embedding | 72.74% |
| 5 | ELMO | 71.81% |
| 6 | FlairEmbeddings (Multi) | 68.28% |
| 7 | **MuRIL** | **80.69%** |
| 8 | XLM-R | 69.42% |

Figure 1: Confusion matrix of the NER model with performance 80.69%

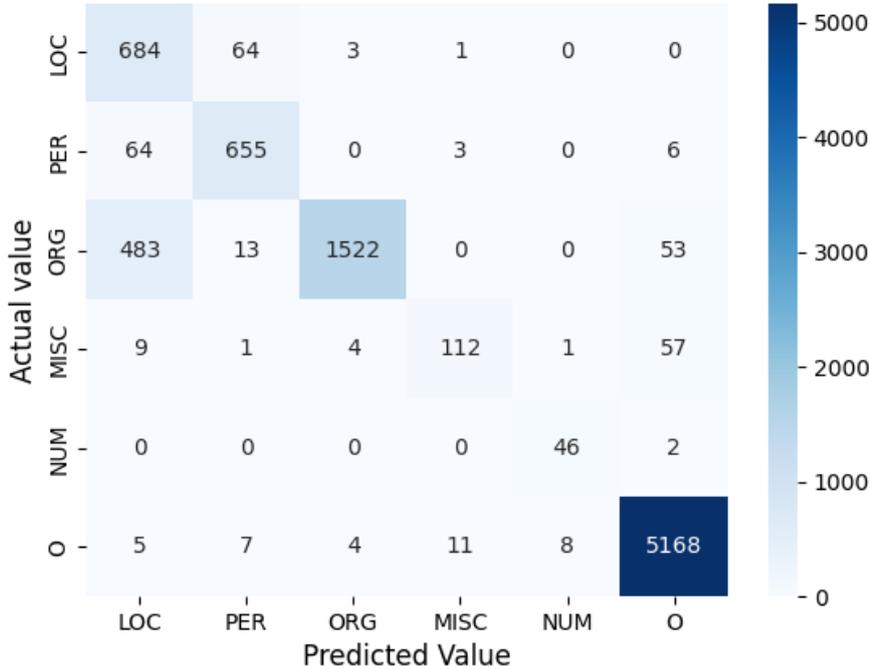

the same training, validation and testing dataset described in the table 2 in all the experiments.

This helps us evaluate the performance of these word embeddings on the AsNER dataset. The micro average score of each tagging model is presented in the table 4. Out of eight different configurations, the NER model that trained using MuRIL (Khanuja et al., 2021) word embedding achieves the highest F1-score of 80.69% among all the models. The FastText, FlairEmbedding and XLM-R show similar performance accuracy of 67.94%, 68.28% and 69.42%, respectively. Whereas the BytePairEmbeddings, BERT and ELMO embedding achieve comparatively higher tagging performance of 76.24%, 72.74% and 71.81%, respectively. The Word2vec reports the lowest performance of 60.23%. Figure 1 shows the confusion matrix of the best model. As there are no existing Assamese NER using the neural model, therefore the best performing model can be considered as a baseline model for Assamese NER.

We observe that ORG is predicted as LOC in numerous instances. PER is more often predicated as LOC. The model is also often confused with ORG, where it tags as "O" in the case of ORG.

## 7. Challenges in Assamese Named Entity Recognition

Assamese is free order language and contains a vast number of polysemous words bearing different meanings. The same word changes meaning according to its grammatical positions in different sentences. In this section, we discuss various challenges in developing NER for Assamese AsNER.

**Lack of resources**: Linguistic resources, such as annotated data, gazetteers, and existing baselines, play a significant role in the named entity recognition process. Such resources are made significant progress for various languages. However, linguistics resources for Assamese are yet to either be developed or mature.

**Ambiguity**: Most of Indian languages contain a large number of polysemous words. They convey different meanings according to their positions in a sentence. For example, মানস /manɔs/ is a name of boy tagged as PERSON. It is also a name of a river and a national park located in Assam tagged as LOCATION. It also means the name of the sacred lake (/manɔs sɔrovɔr/) located in *Kailash mountain* which is also a LOCATION. In some cases, মানস is used as an ADJECTIVE in the Assamese to convey *a desire* or *a wish* or sometimes *mind or conscience*.

**Absence of Capitalization**: Capitalization for a noun is an important feature of a language in recognizing a named entity (NE). It helps to enhance the accuracy of NER system. Unlike English, there is no distinction between plain dictionary words and NEs in Assamese. It makes the named entity recognition process difficult. Examples are সৌৰভ, কলকাতা, আইআইটি গুৱাহাটী (Saurabh, Kolkata, IIT Guwahati). In Assamese, as opposed to English, there is no capitalized concept in the case of nouns. Therefore, a named entity may be missed by annotators and NE recognizers.

**Agglutinative nature**: Agglutination adds additional features in the root word to convey different meaning e.g. ভাৰত /bʰarɔt/ refers to the country India whereas the word ভাৰতীয় /bʰarɔtiyɔ/ refers to the people who are from India.

**Nested entities**: Sometimes, NER tagging conflict occurs in detecting a named entity class when a compound word consists of named entities of different classes. In case of দিল্লী বিশ্ববিদ্যালয় /delhi viswɔvidyalɔy/ (Delhi University) is an organization (ORG), but *Delhi* refers to a location (LOC). Thus it becomes difficult for the recognizer to tag an appropriately named entity class.

## 8. Conclusion

We presented the AsNER dataset and baseline NER for low-resourced, morphologically rich Assamese language. Our contributions can be described in three parts. Firstly, the development of AsNER dataset for Assamese. Secondly, we evaluated AsNER dataset using state-of-the-art sequence tagging architecture by training several NER models using different word embeddings such as Word2Vec (IndicBert), Fasttext, BytePair, BERT, ELMO, MuRIL, XLM-R and FlairEmbedding. Thirdly, we compared the performance of different Assamese NER models trained on our dataset. We found that the model with MuRIL embedding yields the highest accuracy in NER tagging with an F1-score of 80.69%. Conclusively, we discussed various challenges encountered during the annotation and evaluation process.

We believe, our dataset will be a potential resource for the technological development of Assamese language. Our proposed baseline NER model is the first of its kind for Assamese which is developed using deep learning approach. We made the dataset and top performing NER model publicly available. The tagging accuracy is still comparably less than the state-of-the-art NER tagging results. There may be two reasons- a) The POS training size is still not enough for training. b) The language models are still lacking to get the linguistic features of the morphologically rich, highly inflectional language.

## 9. Bibliographical References

Akbik, A., Blythe, D., and Vollgraf, R. (2018). Contextual string embeddings for sequence labeling. In *Proceedings of the 27th international conference on computational linguistics*, pages 1638–1649.

Akbik, A., Bergmann, T., Blythe, D., Rasul, K., Schweter, S., and Vollgraf, R. (2019). Flair: An easy-to-use framework for state-of-the-art nlp. In *Proceedings of the 2019 Conference of the North American Chapter of the Association for Computational Linguistics (Demonstrations)*, pages 54–59.

Babych, B. and Hartley, A. (2003). Improving machine translation quality with automatic named entity recognition. In *Proceedings of the 7th International EAMT workshop on MT and other language technology tools, Improving MT through other language technology tools, Resource and tools for building MT at EACL 2003*.

Bojanowski, P., Grave, E., Joulin, A., and Mikolov, T. (2017). Enriching word vectors with subword information. *Transactions of the Association for Computational Linguistics*, 5:135–146.

Chinchor, N. and Robinson, P. (1997). Muc-7 named entity task definition. In *Proceedings of the 7th Conference on Message Understanding*, volume 29, pages 1–21.

Chiu, J. P. and Nichols, E. (2016). Named entity recognition with bidirectional lstm-cnns. *Transactions of the Association for Computational Linguistics*, 4:357–370.

Conneau, A., Khandelwal, K., Goyal, N., Chaudhary, V., Wenzek, G., Guzmán, F., Grave, E., Ott, M., Zettlemoyer, L., and Stoyanov, V. (2019). Unsupervised cross-lingual representation learning at scale. *arXiv preprint arXiv:1911.02116*.

Devlin, J., Chang, M.-W., Lee, K., and Toutanova, K. (2018). Bert: Pre-training of deep bidirectional transformers for language understanding. *arXiv preprint arXiv:1810.04805*.

Etzioni, O., Cafarella, M., Downey, D., Popescu, A.-M., Shaked, T., Soderland, S., Weld, D. S., and Yates, A. (2005). Unsupervised named-entity extraction from the web: An experimental study. *Artificial intelligence*, 165(1):91–134.

Heinzerling, B. and Strube, M. (2018). BPEmb: Tokenization-free Pre-trained Subword Embeddings in 275 Languages. In Nicoletta Calzolari (Conference chair), et al., editors, *Proceedings of the Eleventh International Conference on Language Resources and Evaluation (LREC 2018)*, Miyazaki, Japan, May 7-12, 2018. European Language Resources Association (ELRA).

Huang, Z., Xu, W., and Yu, K. (2015). Bidirectional lstm-crf models for sequence tagging. *arXiv preprint arXiv:1508.01991*.

Hvingelby, R., Pauli, A. B., Barrett, M., Rosted, C., Lidegaard, L. M., and Søgaard, A. (2020). Dane: A named entity resource for danish. In *Proceedings of the 12th Language Resources and Evaluation Conference*, pages 4597–4604.

Kakwani, D., Kunchukuttan, A., Golla, S., Gokul, N., Bhattacharyya, A., Khapra, M. M., and Kumar, P. (2020). inlpsuite: Monolingual corpora, evaluation benchmarks and pre-trained multilingual language models for indian languages. In *Proceedings of the 2020 Conference on Empirical Methods in Natural Language Processing: Findings*, pages 4948–4961.

Khanuja, S., Bansal, D., Mehtani, S., Khosla, S., Dey, A., Gopalan, B., Margam, D. K., Aggarwal, P., Nagipogu, R. T., Dave, S., Gupta, S., Gali, S. C. B., Subramanian, V., and Talukdar, P. (2021). Muril: Multilingual representations for indian languages.

Lample, G., Ballesteros, M., Subramanian, S., Kawakami, K., and Dyer, C. (2016). Neural architectures for named entity recognition. *arXiv preprint arXiv:1603.01360*.

Larsen, B. (1999). A trainable summarizer with knowledge acquired from robust nlp techniques. *Advances in automatic text summarization*, 71.

Mikolov, T., Chen, K., Corrado, G., and Dean, J. (2013). Efficient estimation of word representations in vector space. *arXiv preprint arXiv:1301.3781*.

Mollá, D., Van Zaanen, M., Smith, D., et al. (2006). Named entity recognition for question answering.

Neudecker, C. (2016). An open corpus for named entity recognition in historic newspapers. In *Proceedings of the Tenth International Conference on Language Resources and Evaluation (LREC'16)*, pages 4348–4352.

Pan, X., Zhang, B., May, J., Nothman, J., Knight, K., and Ji, H. (2017). Cross-lingual name tagging and linking for 282 languages. In *Proceedings of the 55th Annual Meeting of the Association for Computational Linguistics (Volume 1: Long Papers)*, pages 1946–1958.

Peters, M. E., Neumann, M., Iyyer, M., Gardner, M., Clark, C., Lee, K., and Zettlemoyer, L. (2018). Deep contextualized word representations. *arXiv preprint arXiv:1802.05365*.

Sang, E. F. and De Meulder, F. (2003). Introduction to the conll-2003 shared task: Language-independent named entity recognition. *arXiv preprint cs/0306050*.

Strötgen, J. and Gertz, M. (2013). Multilingual and cross-domain temporal tagging. *Language Resources and Evaluation*, 47(2):269–298.

Sutskever, I., Vinyals, O., and Le, Q. V. (2014). Sequence to sequence learning with neural networks. *arXiv preprint arXiv:1409.3215*.

Yu, J., Bohnet, B., and Poesio, M. (2020). Named entity recognition as dependency parsing. *arXiv preprint arXiv:2005.07150*.

Zhang, Z., Han, X., Liu, Z., Jiang, X., Sun, M., and Liu, Q. (2019). Ernie: Enhanced language representation with informative entities. *arXiv preprint arXiv:1905.07129*.